%% file: paper.tex
\documentclass[sigconf]{acmart}

\AtBeginDocument{%
  \providecommand\BibTeX{{%
    \normalfont B\kern-0.5em{\scshape i\kern-0.25em b}\kern-0.8em\TeX}}}

\setcopyright{acmcopyright}
\copyrightyear{2018}
\acmYear{2018}
\acmDOI{XXXXXXX.XXXXXXX}

\acmConference[BIOKDD '23]{Proceedings of
22nd International Workshop on Data Mining in Bioinformatics}{August 09, 2023}{Long Beach, CA}
\acmPrice{15.00}
\acmISBN{978-1-4503-XXXX-X/18/06}

\title{Precision Anti-Cancer Drug Selection via Neural Ranking}
\author{Vishal Dey}
\affiliation{
\institution{The Ohio State University}
\city{Columbus}
\country{USA}}
\email{dey.78@osu.edu}

\author{Xia Ning}
\authornote{Corresponding Author}
\affiliation{
\institution{The Ohio State University}
\city{Columbus}
\country{USA}
}
\email{ning.104@osu.edu}

\usepackage{amsthm,amsmath}
\usepackage{chemformula} 
\usepackage[T1]{fontenc} 
\usepackage{amsmath}
\usepackage{bbm}
\usepackage{subcaption,float,placeins}
\usepackage[flushleft]{threeparttable}
\usepackage{multirow,booktabs}
\usepackage{xspace}
\usepackage{verbatim} 
\usepackage{comment, mathtools}
\usepackage{longtable, threeparttablex}
\usepackage{graphicx}
\usepackage{soul}
\usepackage{siunitx}
\usepackage{url}
\usepackage[colorinlistoftodos]{todonotes}
\usepackage{subcaption}
\usepackage[utf8]{inputenc} 
\usepackage{geometry}

\include{define}

\begin{document}

\begin{abstract}
Personalized cancer treatment requires a thorough understanding of 
complex interactions between drugs and cancer cell lines in varying genetic and molecular contexts.
To address this, high-throughput screening has been used to generate
large-scale drug response data, facilitating data-driven computational models.
Such models can capture complex drug-cell line interactions across various contexts in a fully data-driven manner.
However, accurately prioritizing the most sensitive drugs for each cell line still remains a significant challenge.
To address this, we developed neural ranking approaches
that leverage large-scale drug response data across multiple cell lines from diverse cancer types.
Unlike existing approaches that primarily utilize regression and classification techniques for drug response prediction,
we formulated the objective of drug selection and prioritization as a drug ranking problem.
In this work, we proposed two neural listwise ranking methods that learn latent representations
of drugs and cell lines, and then use those representations to score drugs in each cell line via
a learnable scoring function.
Specifically, we developed a neural listwise ranking method, \ListNet,
on top of the existing method ListNet.
Additionally, we proposed a novel listwise ranking method, \SelNet,
that focuses on all the sensitive drugs instead of the top sensitive drug, unlike \ListNet.
%
Our results demonstrate that \SelNet outperforms the best baseline
with significant improvements of as much as 8.6\% in hit@20 across 50\% test cell lines.
Furthermore, our analyses suggest that the learned latent spaces from our proposed methods
demonstrate informative clustering structures and capture
relevant underlying biological features. 
Moreover, our comprehensive empirical evaluation provides a thorough and objective comparison of the performance of different methods (including our proposed ones).
\end{abstract}


\maketitle


\section{Introduction}
\label{sec:intro}
Precision cancer treatment aims to tailor therapies to individual patients
by identifying effective anti-cancer drugs for each patient
based on their unique genetic and molecular characteristics.
However, this requires an in-depth understanding of complex interactions between drugs and patients in 
different genetic and molecular contexts, which presents a significant challenge.
To address this challenge, high-throughput screening methods\cite{Gupta2009} have been used
to generate large-scale drug response data\cite{Yang2012,Barretina2012},
providing a valuable resource for developing data-driven models.
%
However, developing models that can accurately
encode complex interactions between cell lines and drugs
and prioritize the most promising anti-cancer drugs for each cell line still remains a challenging task.
To tackle these challenges, data-driven models jointly utilize drug response data across
multiple cell lines and cancer types, motivated by findings from pan-cancer studies\cite{Weinstein2013}.
These studies suggest that there are commonalities in genetic and molecular features among different cancer types\cite{Kandoth2013}.
Following such insights, in this paper, we develop computational approaches
that can leverage drug response data across a large number of cell lines from diverse cancer types
to identify and prioritize anti-cancer drugs in each cell line.
Unlike existing computational approaches, our approach is inspired by learning-to-rank (LeToR) methods\cite{Burges2005},
which can naturally formulate the objective of anti-cancer drug selection and prioritization.

In this work, we develop neural listwise LeToR methods
that learn latent representations of cell lines and drugs in a data-driven manner, and
then use those representations to score drugs in each cell line
via a learnable scoring function.
%
%
We utilize the existing listwise ranking objective\cite{Cao2007} to develop a
neural listwise ranking method, denoted as \ListNet, which
considers the entire ranking structure at a time
and learns the probability of the most sensitive drugs in each cell line
being ranked at the top (i.e., top-one probability distribution).
By minimizing the discrepancy between the predicted and ground-truth top-one probability distributions,
%
\ListNet learns to appropriately score the drugs
leading to an accurate selection of the most sensitive drug in each cell line.

In addition, 
we propose another listwise ranking method for drug selection, denoted as \SelNet, that focuses on
selecting all the sensitive drugs. 
%
%
Additionally, we evaluated our proposed methods against strong regression and pairwise ranking baselines.
Our results demonstrate that \SelNet mostly outperforms the best baselines
with significant improvements of much as 8.6\% in hit@20 across 50\% test cell lines.
Furthermore, our analyses suggest that the learned latent spaces from our proposed methods
demonstrate informative clustering structures and capture
relevant underlying biological features (e.g., cancer types, drug mechanism of action).
%

The rest of the manuscript is organized as follows. Section~\ref{sec:rel} presents the 
related work on computational methods in anti-cancer drug response prediction and drug prioritization.
Section~\ref{sec:methods} presents the proposed listwise methods and
Section~\ref{sec:materials} describes the datasets, baseline methods, experimental settings and evaluation metrics.
Section~\ref{sec:results} presents an overall comparison of all methods in one experimental setting across both datasets
and detailed analyses of embeddings.
Section~\ref{sec:conclusion} concludes the paper.

\section{Related Works}
\label{sec:rel}

\subsection{Computational Methods in Drug Response Prediction}
With an increasing abundance of large-scale drug response data and 
advanced high-throughput screening\cite{Gupta2009}, 
data-driven computational approaches have been developed
for drug response prediction in cancer cell lines. 
Following pan-cancer studies\cite{Weinstein2013}, these approaches
have been extended beyond singe-drug or single-cell line modeling
to jointly leverage the drug response data
across multiple drugs and cell lines.
This enables such approaches to capture the interactions
among multiple drugs, among multiple cell lines, and between drugs and cell lines.
%
Typically, these approaches either focus on regression\cite{Su2019}
which estimates the drug responses for a given cell line,
or on classification\cite{Bonanni2022} which predicts whether a drug is sensitive or not in a given cell line.  
These approaches employ
various machine learning techniques such as kernel methods\cite{He2018KRL},
matrix factorization\cite{Wang2017}, 
and deep learning\cite{Baptista2020,Zuo2021}.
%
We refer the readers to a comprehensive survey\cite{Firoozbakht2022}
for broader coverage of the existing literature in this area.
In contrast to the most popular approaches toward drug response prediction,
our work is more related to LeToR approaches since it naturally
models drug selection and prioritization.

\subsection{LeToR methods in Drug Prioritization}
Unlike the aforementioned regression and classification methods,
LeToR methods for drug prioritization
are relatively under-explored
\cite{He2018KRL,He2018,Gerdes2021,Prasse2022}.
LeToR methods focus on learning to appropriately score the candidate drugs and to optimize different
objectives so as to achieve accurate ranking.
LeToR methods can be broadly categorized into three approaches:
pointwise\cite{Burges2005}, pairwise\cite{Burges2005} and listwise\cite{Cao2007}.
%
In fact, the pointwise approach 
typically performs inferior to both pairwise and listwise approaches\cite{Cao2007} since
the ranking structure is not explicitly leveraged.
%
One of the popular pairwise ranking approaches for drug prioritization, \cclerank{\cite{He2018}},
do not explicitly leverage auxiliary information such as molecular structures,
which are known to be well correlated to activity\cite{Gillet1998},
drug-likeliness\cite{Bickerton2012}, 
and other pharmacological properties\cite{Lipinski2001}.
This may hinder such models to learn the above-mentioned structure-activity correlations, 
a key to many aspects in drug discovery\cite{Perkins2003}. 

In addition to pairwise approaches, listwise approaches have been utilized in recent works.
Kernelized Rank Learning (KRL)\cite{He2018KRL} is a listwise LeToR method that
optimizes an upper bound of the Normalized Discounted Cumulative Gain (NDCG@k),
and learns to approximate the drug sensitivities via a kernelized linear regression. 
However, KRL notably underperforms \cclerank
across multiple experimental settings as demonstrated by He et al\cite{He2018}.
%
Another neural listwise ranking method developed by Prasse et al.\cite{Prasse2022}
optimizes a smooth approximation of NDCG@k.
However, the experiments from this study are not adequately comprehensive and may
not be directly comparable to other studies in the literature
due to their usage of multi-omics profiles and customized definitions of ground-truth drug relevance scores,
which deviates from the standard approach in other studies.
Additionally, the proposed method was not evaluated against state-of-the-art pointwise or pairwise approaches.
Furthermore, the experiments were limited to one experimental setting (`Cell cold-start'), which may restrict
the generalizability of their findings.
%

\section{Methods}
\label{sec:methods}
\input{tables/notation}
Table~\ref{tbl:notations} presents the key notations used in the manuscript.
Drugs are indexed by $i$ and $j$ in the set of drugs \SetDrugs, and cell lines are denoted by $c \in \SetCells$.
In this manuscript, $\SetDrugs^{+}_c$ / $\SetDrugs^{-}_c$ indicate
the set of sensitive and insensitive drugs, respectively, in the cell line $c$.
For example, $d_i \in \SetDrugs^{+}_c$ denotes a sensitive drug $i$ in cell line $c$;
$d_j \in \SetDrugs^{-}_c$ denotes an insensitive drug $j$ in cell line $c$.
In this section, we proposed 
two listwise learning-to-rank methods (\ListNet and \SelNet) for anti-cancer drug selection and prioritization.
We first introduce the overall architecture of our methods in Section~\ref{sec:methods:overall},
and then discuss each component in detail in Sections~\ref{sec:methods:clv} and \ref{sec:methods:dv}.
We discuss each of our proposed methods and their ranking optimization process in subsequent sections.

\subsection{Overall Framework}
\label{sec:methods:overall}
\begin{figure*}[h!]
\includegraphics[width=0.9\textwidth]{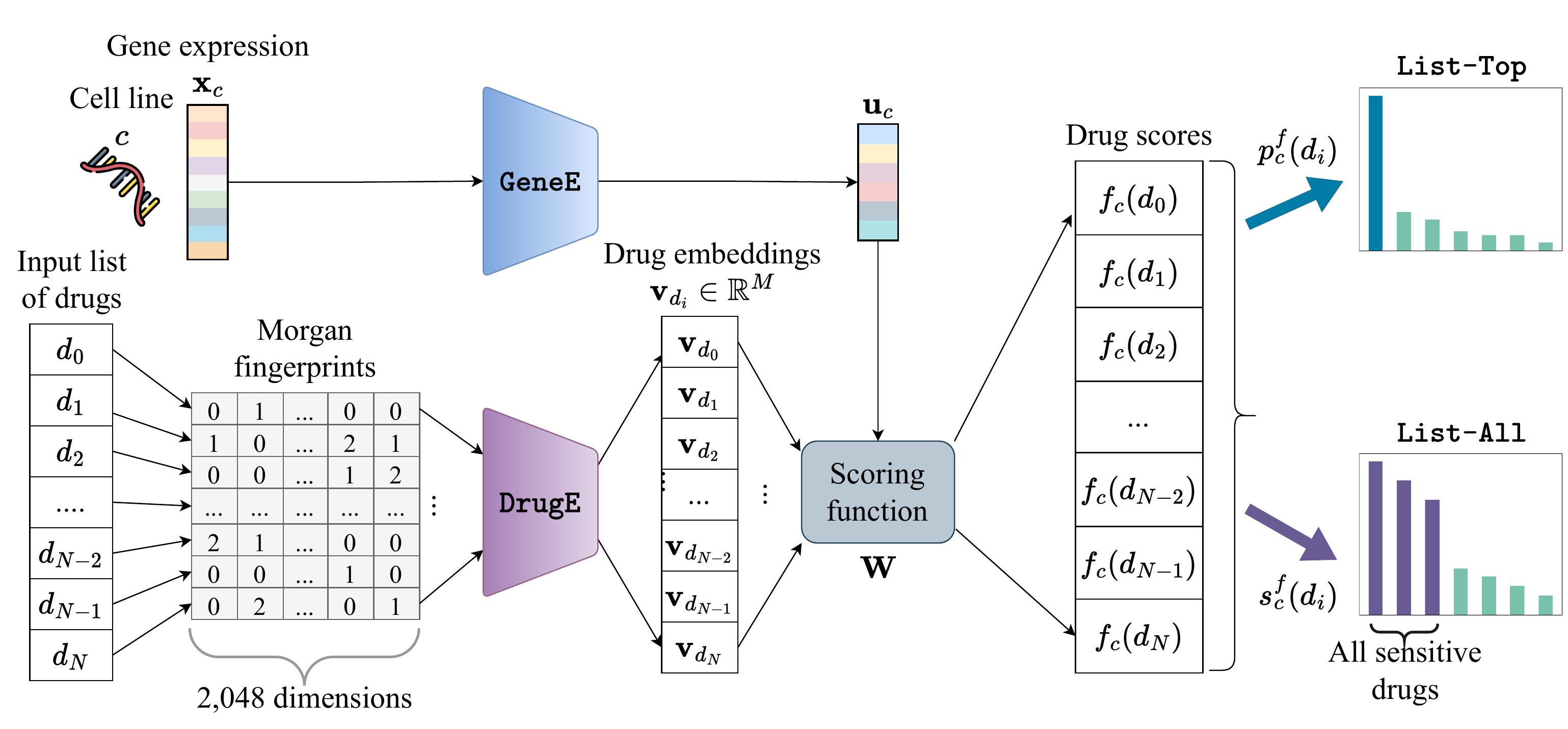}
\caption{Overall Framework.
The cell line embeddings $\CellEmb_c$ and drug embeddings $\DrugEmb_{d_i}$ are used to score the drugs $d_i \in \SetDrugs_c$. 
Each ranking method utilizes a different ranking objective and thus utilizes the scores $f_c(d_i)$ differently.
The pretrained encoder \GeneE is finetuned during ranking optimization.
}
\label{fig:framework}
\end{figure*}

In order to select and prioritize sensitive drugs in each cell line,
our proposed LeToR methods optimize different objectives that 
inducing the correct ranking structure among the top-sensitive or all involved drugs in each cell line.
%
Figure~\ref{fig:framework} presents an overall scheme of our proposed methods.
To induce the correct ranking structure, each method learns to
accurately score drugs in each cell line
using the learned cell line and drug embeddings.
The embeddings and scoring function are learned
in a fully data-driven manner from the drug response data.
%
Intuitively, the cell line latent space embeds the genomic and response information of cell lines,
while the drug latent space embeds the structural and sensitivity information for drugs.
The cell line embeddings are initially learned from the gene expression profiles
using a pre-trained auto-encoder model \GeneAE (Section~\ref{sec:methods:clv}).
The drug embeddings are learned from the molecular fingerprints (Section~\ref{sec:methods:dv}).
During training, the cell line and drug embeddings are then used and updated 
to correctly score drugs against each cell line
using a learnable scoring function (Section~\ref{sec:methods:lrank:score}). 
%
Note that \ListNet and \SelNet
utilize the same scoring function, however, optimize separate ranking objectives.
%

\subsection{Pretraining for Cell Line Embeddings}
\label{sec:methods:clv}

\begin{figure}[h!]
\includegraphics[width=0.5\textwidth, trim={0cm 2.1cm 0.5cm 0.25cm}, clip]{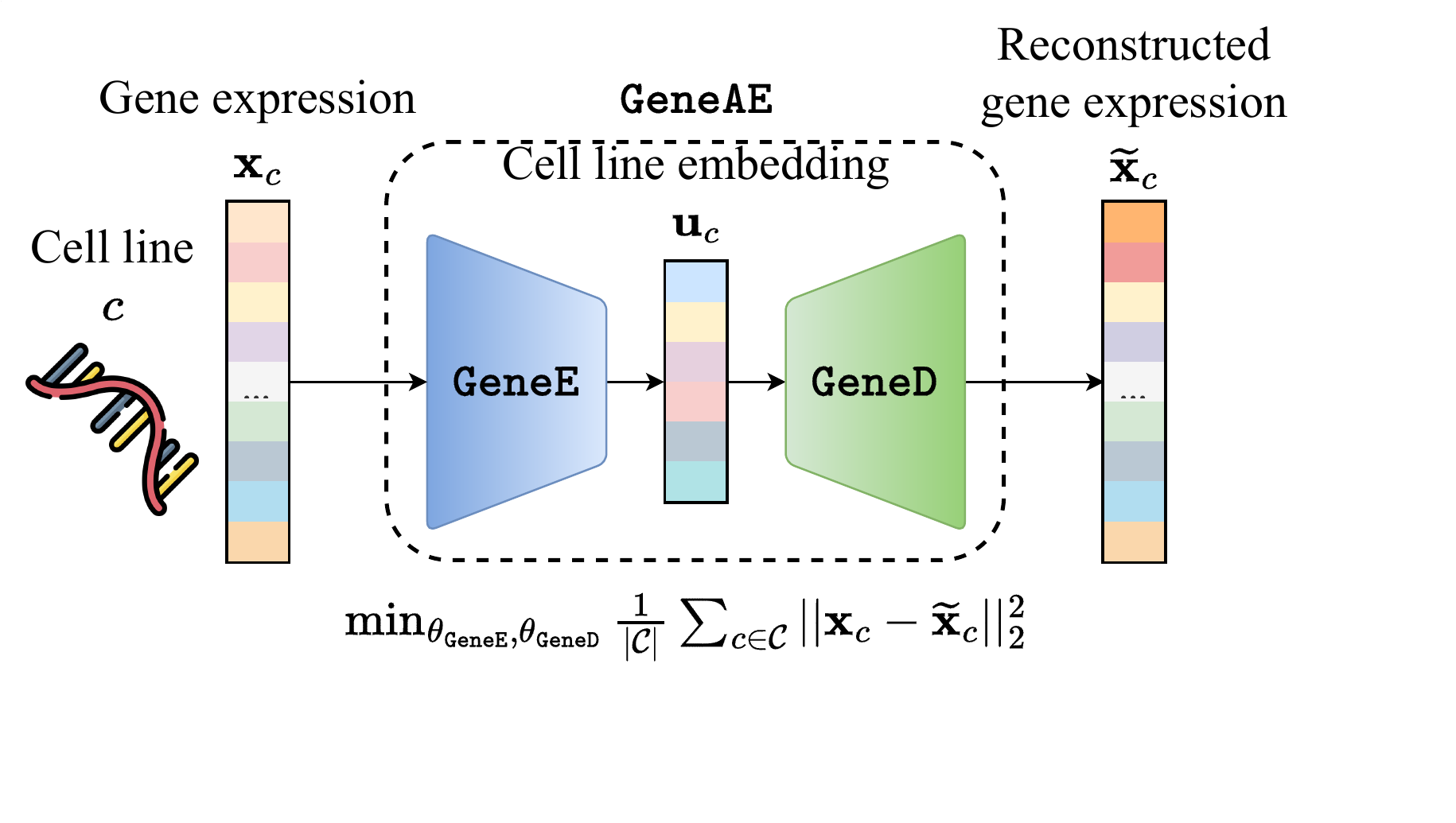}
\caption{Pretraining framework.
Given a gene expression $\mathbf{x}_c$, \GeneAE reconstructs it as $\widetilde{\mathbf{x}}_c$ 
through the auto-encoder, meanwhile learning an embedding $\CellEmb_c$ in the latent space.
}
\label{fig:geneae}
\end{figure}

%
In order to learn rich informative cell line embeddings,
we pretrain a stacked auto-encoder framework \GeneAE, 
similar to existing gene expression auto-encoder frameworks{\cite{Xie2017,Dincer2020}}.
Figure~\ref{fig:geneae} presents the architecture of \GeneAE. 
%
\GeneAE embeds the rich genomic information into a latent space,
and learns the complex and non-linear interactions among genes.
Specifically, \GeneAE leverages the gene expression profile $\mathbf{x}_c$ to learn a
low-dimensional embedding $\CellEmb_c$ via the encoder \GeneE,
followed by reconstruction of the expression profile from $\CellEmb_c$ via the decoder \GeneD.
These embeddings out of the pretrained \GeneE are used to score
drugs in each cell line during the downstream ranking (Section~\ref{sec:methods:lrank:score}).
Such embeddings can be utilized as transferable representations of cell lines
that can potentially
enable better generalizability of downstream drug scoring/ranking models.
In summary, these embeddings can potentially improve the performance of drug ranking models
by leveraging the shared biological features across cell lines.

As any other pretraining module\cite{Erhan2010}, 
\GeneAE has two training stages: pretraining, and finetuning.
During pretraining, the model parameters are learned via back-propagation by minimizing the 
reconstruction error (here, MSE) from the actual and reconstructed gene expression data as,
\begin{equation}
\label{equ:mse}
\min\nolimits_{\theta_{\textsc{\GeneE}}, \theta_{\textsc{\GeneD}}} \frac{1}{|\SetCells|} \sum_{c \in \SetCells} || \mathbf{x}_c - \widetilde{\mathbf{x}}_c||_2^2,
\end{equation}
where $\mathbf{x}_c$ denotes the input gene expression of cell line $c$,
 $\widetilde{\mathbf{x}}_c = \GeneD (\GeneE (\mathbf{x}_c))$ denotes the corresponding reconstructed gene expression,
 $\theta_\textsc{\GeneE}$ and $\theta_\textsc{\GeneD}$ denote the learnable parameters of \GeneE and \GeneD, respectively,
and \SetCells denotes the set of input cell lines.
%
During pretraining, parameters of both \GeneE and \GeneD modules are learned, and the
learned parameters of \GeneE are transferred and finetuned during the optimization for downstream ranking tasks.
The finetuning of pretrained \GeneE adapts the output embeddings
toward the specific downstream ranking. 

\subsection{Embedding Drugs from Fingerprints}
\label{sec:methods:dv}

In this work, molecular fingerprints\cite{Rogers2010} are leveraged to learn
informative drug embeddings via a low-dimensional projection.
These fingerprints are discrete feature vectors representing the presence of
molecular substructures in a drug given a fixed vocabulary.
%
Typically, such fingerprints need to be high-dimensional to sufficiently
capture all relevant structural information.
On the other hand, the drug embeddings 
can selectively encode relevant structural
information (specific to the ranking task) as non-linear functions of input fingerprints.
%
These embeddings are further used to score drugs in each cell line,
and are learned during ranking optimization given drug response data across multiple cell lines.
Such learned embeddings enable accurate drug scoring since
similar drugs in terms of structures and sensitivities across multiple cell lines
obtain similar embeddings.
%
To learn such embeddings of dimension $M$ ($M$ is a hyperparameter),
we used a fully connected neural network \DrugE.
As inputs to \DrugE, we used Morgan count fingerprints\cite{Rogers2010} with radius = 3 and
2,048 bits.
%
While graph neural networks (GNNs)\cite{Wu2021} have demonstrated promising empirical performance
in molecular prediction tasks\cite{Zhang2022},
we observed inferior ranking performance from the drug embeddings learned from GNNs
compared to those learned from fingerprints, from preliminary experiments.
This is possibly due to the limited number of unique drugs in our datasets.

\subsection{Listwise Ranking for Top-One Drug: \ListNet}
\label{sec:methods:lrank}

We adopted the standard ListNet\cite{Cao2007} objective
to develop a neural listwise ranking method, \ListNet.
%
%
\ListNet considers the entire ranking structure at a time
and focuses on accurately estimating the top-one probability of drugs.
The top-one probability of a drug $d$, denoted as $p_{c}(d)$, 
is its probability of being ranked at the top
given the scores of all involved drugs in the cell line $c$.
Formally, the predicted top-one probability denoted as $p_{c}^{f}(d)$ is defined as follows:
\begin{equation}
\label{equ:ptop}
p_{c}^{f}(d) = \frac{\exp{(f_c(d))}}{\sum_{d_j \in \SetDrugs_{c}} \exp{(f_c(d))}},
\end{equation}
where $f_c(d)$ denotes the score of drug $d$ in cell line $c$ (which will be discussed later in Section~\ref{sec:methods:lrank:score}).
The top-one probabilities are optimized using the cross-entropy loss as follows:
\begin{equation}
\label{equ:opt2}
\min_{\Theta} - \sum_{c \in \SetCells} \left[ \sum_{d \in \SetDrugs_{c}}  p_c^{top}(d) \log(p_{c}^{f}(d)) \right] ,
\end{equation}
where $\Theta$ denotes the learnable parameters in the model;
and $p_c^{top}(d)$ denotes the ground-truth top-one probability of drug $d$ in the cell line $c$ 
according to the ground-truth drug responses.
In this study, the drug responses are quantified with \underline{A}rea \underline{U}nder the dose-response \underline{C}urve denoted as AUC,
where smaller AUC values indicate higher drug sensitivities.
$p_c^{top}(d)$ is computed via Equation~\ref{equ:ptop} by replacing $f_c(d)$ with the negated AUCs
(since lower AUCs indicate higher drug sensitivities).
%
Minimizing the above loss reduces the discrepancy between the
predicted and the ground-truth top-one probability distribution over drugs in each cell line.
This results in an accurate estimation of the top-one probability,
which enables an accurate selection of the most sensitive drug in each cell line.
During the optimization, \ListNet (and \SelNet presented in the following section similarly)
finetunes \GeneE (Section~\ref{sec:methods:clv})
and learns \DrugE (Section~\ref{sec:methods:dv}).
This enables the cell line and drug embeddings out of \GeneE and \DrugE, respectively, to encode 
task-relevant information.

\subsubsection{Drug Scoring}
\label{sec:methods:lrank:score}
%
%
In order to score the drug $d$ in a cell line $c$,
we used a parameterized bilinear function denoted as $f_c(d)$.
The function $f_c(d)$ is parameterized via a learnable weight matrix 
$\mathbf{W} \in \mathbb{R}^{|\CellEmb_c| \times |\DrugEmb_d|}$,
and is applied over the $\CellEmb_c \in \mathbb{R}^{|\CellEmb_c|}$ 
and $\DrugEmb_d \in \mathbb{R}^{|\DrugEmb_d|}$ as follows:
\begin{equation}
\label{equ:score}
f_c(d) = \CellEmb_c^\intercal \mathbf{W} \DrugEmb_d,
\end{equation}
where, $\mathbf{W}$ is learned via backpropagation in an end-to-end manner during optimization.
%
Intuitively, 
the learnable full-rank weight matrix $\mathbf{W}$ in the bilinear scoring function
can capture complex and relevant interactions between the two latent vectors.
%
Once the scores for all drugs in a cell line $c$ (i.e., $\SetDrugs_c$) are obtained,
the drugs $\SetDrugs_c$ are sorted based on such scores in descending order.
The most sensitive drugs in $c$ will have higher scores than the insensitive ones in $c$.
%
Note that ranking-based methods such as our proposed ones
will achieve optimal ranking performance as long as
these scores induce correct ranking structure;
the scores do not need to be exactly identical to the drug response scores (i.e., AUC values).
%

\subsection{Listwise Ranking for All Sensitive Drugs: \SelNet}
\label{sec:methods:srank}

Since \ListNet focuses solely on the top-ranked drug, it may lead to
suboptimal performance in terms of selecting all the sensitive drugs in each cell line,
as demonstrated in our experiments (Section \ref{sec:results:lco}),
To address this, we proposed a new listwise neural ranking method, \SelNet
with an objective that can optimize the selection of all sensitive drugs in each cell line.
\SelNet leverages the entire ranking structure at a time
and follows a similar architecture to \ListNet.
But unlike \ListNet, 
\SelNet estimates the probability of a drug being sensitive given the scores of all drugs,
where higher scores induce higher probabilities.
Since the estimated probability of each drug being sensitive is dependent on the scores of all other drugs
in the list, \SelNet is a listwise ranking method.
\SelNet aims to minimize the distance between such estimated score-induced probabilities 
and the ground-truth sensitivity labels across all drugs in each cell line.
%
%
Specifically, \SelNet is trained by minimizing the following loss:
\begin{equation}
\label{equ:opt3}
\min_{\Theta} - \sum_{c \in \SetCells} \left[ \sum_{d \in \SetDrugs_{c}}  l_c(d) \log(s_{c}^{f}(d)) \right] ,
\end{equation}
where $\Theta$ denotes the learnable parameters in the model;
$l_c(d)$ is a binary sensitivity label indicating whether drug $d$ is sensitive in cell line $c$;
and $s_c^{f}(d)$ denotes the probability of drug $d$ to be sensitive in the cell line $c$.
Formally, $s_c^{f}(d)$ is computed from the predicted scores via the parameterized softmax as:
\begin{equation}
\label{equ:ps}
s_{c}^{f}(d) = \frac{\exp{(f_c(d)/\tau)}}{\sum_{d_j \in \SetDrugs_{c}} \exp{(f_c(d)/\tau)}},
\end{equation}
where $\tau$ is the temperature (a scaling factor > 0) that controls the softness/sharpness of the score-induced probability distribution
while maintaining the relative ranks.
A lower scaling factor results in a sharper probability distribution with higher probabilities on very few drugs.
Note that the scaling factor can also be applied similarly in Equation~\ref{equ:ptop}, however, we observed no notable performance
difference empirically. For \SelNet, we fix $\tau$ to 0.5.
Note that the optimization objective (Equation~\ref{equ:opt3}) resembles the ListNet objective (Equation~\ref{equ:opt2})
in the sense that both
aim to minimize the cross-entropy between two score-induced empirical probability distributions. 

\section{Materials}
\label{sec:materials}

In this section, we present the data sets and baselines used in Sections~\ref{sec:materials:data} and \ref{sec:materials:baselines},
respectively;
the experimental setting in Section~\ref{sec:materials:expts}; and
the evaluation metrics adopted to evaluate ranking performance in Section~\ref{sec:materials:eval}.

\subsection{Dataset}
\label{sec:materials:data}
\vspace{-10pt}
\input{tables/data}
\vspace{-5pt}

We collected the drug response data set from the
Cancer Therapeutic Response Portal version 2
(CTRP)\footnote{\url{https://ctd2-data.nci.nih.gov/Public/Broad/CTRPv2.0_2015_ctd2_ExpandedDataset/} (accessed on 01/20/22)}\cite{Seashore2015}.
%
We focused on this data set
because it covers a large number of
cell lines and drugs compared to other available data sets\footnote{Due to space limitations, we present the results only on one dataset in this workshop paper.
Additional results for other experimental settings and datasets will be published in a forthcoming full-paper version.}.
We utilized the Cancer Cell Line Encyclopedia version 22Q1
(CCLE)\footnote{\url{https://depmap.org/portal/download/} (accessed on 01/20/22)}\cite{Barretina2012}
for the gene expression data. 
CCLE provides multi-omics data (genomic, transcriptomic and epigenomic)
for more than 1,000 cancer cell lines.
However, in this study, we only used the gene expression (transcriptomic) data following~\cite{He2018}.
%
%
The drug responses are measured using AUC sensitivity scores,
with lower AUC indicating higher sensitivity of a drug in a cell line.
For the drugs with missing responses in a cell line, the corresponding drug-cell line pairs were not included in the training of models.
%
%
For the cell lines that could not be mapped to CCLE, those cell lines and their associated
drug responses were excluded from our experiments.
Since CTRP has more cell lines than other available datasets in the literature,
it is an appropriate choice for evaluating computational methods in
drug selection for new cell lines, 
which is the primary focus of this work.
This work is motivated based on the belief that
such a setup is more relevant to real-life scenarios where the goal is to
suggest potential anti-cancer drugs for new patients.

\subsection{Baselines}
\label{sec:materials:baselines}
We use two strong baseline methods: \cclerank~\cite{He2018}
and \deepcdr~\cite{Liu2020}.
Unlike our proposed methods,
\cclerank is a pairwise ranking approach that learns the drug and cell line embeddings
by explicitly pushing sensitive drugs to the top of the ranking list
and by further optimizing the ranking structure among the sensitives.
%
%
Unlike \cclerank, our proposed methods leverage drug structural and gene expression information 
to learn more informative embeddings 
that may enable improved ranking performance.
Additionally, different from \cclerank, our proposed methods utilize a learnable scoring function
to capture the complex interactions between embeddings.
While \cclerank explicitly enforces similarity regularization on cell line embeddings
using the gene expression-based similarity of cell lines,
our methods enforce such genomic similarity by embedding cell lines
in the latent space via the pretrained \GeneE.
%
Different from our proposed methods and the baseline \cclerank,
\deepcdr, one of the state-of-the-art regression models	
for anti-cancer drug response prediction, 
learns to estimate the exact response scores of every drug in each cell line.

\subsection{Experimental Setting}
\label{sec:materials:expts}
According to the setting of He et al.~\cite{He2018}, 
a percentile labeling scheme was used to label drugs as sensitive or insensitive.
The sensitivity threshold for each cell line 
was determined as the top-5 percentile of its drug responses.
In order to assess the ranking performance on new cell lines,
we employed a leave-cell-lines-out (LCO) validation setting such that
this setting resembles the real-world scenario when known drugs are investigated
for their sensitivity or anti-cancer potential in new patients.
We randomly split all the cell lines from each cancer type into five folds.
In each run, we used the four folds from each cancer type for training
and the other fold for testing.
We use the cell lines from all the cancer types for 4 folds and their corresponding
drug response data collectively for training.
We use the cell lines in the other left-out fold as new (unseen) cell lines for model testing.
This process was repeated five times with each fold serving as the test fold exactly once.
For \cclerank, we follow the `leave-one-out' setup\cite{He2018} in that
the cell line embeddings were learned only for the training cell lines,
and the embeddings for the test cell lines were interpolated
from the nearest neighboring training cell lines in the latent space.
%
For our proposed methods, we use
the gene expression profiles of only the training cell lines to pretrain \GeneAE.
%
%
%

\subsection{Evaluation Metrics}
\label{sec:materials:eval}

In order to evaluate the ranking performance, we generated the true ranking list using the ground-truth AUC response values
and the predicted ranking list using the estimated scores out of the models.
We then compared the two ranking lists (or a portion of them) using popular evaluation metrics:
average precision at $K$ ({\ap}@K), and
average hit at $K$ ({\ah}@K),
which are commonly used in Information Retrieval systems.
%
%
Higher {\ap}@K and {\ah}@K indicate better drug selection where the top-ranked drugs in the predicted ranking list are sensitive.
In addition to \ap and \ah, concordance index (\ci) and concordance index among the sensitive drugs (\sci) \cite{He2018}
are also used
to evaluate the overall quality of the predicted ranking structure among all drugs and sensitive drugs, respectively.
Note that high \ci and \sci values do not necessarily result in high \ap or \ah since
the ranking structure can be well preserved without pushing the few sensitive drugs 
(which constitutes only 5\% of the total drugs) to the very top.
On the other hand, high \ap/\ah indicates the most sensitive drugs are ranked at the top,
but this does not necessarily result in high \ci/\sci. 
In this work, since we primarily focus on identifying the top-$k$ most sensitive drugs in each cell line,
we prioritize and emphasize the \ap and \ah metrics over the \sci and \ci metrics
when evaluating and interpreting our results.

\section{Results}
\label{sec:results}

\subsection{Overall Comparison}
\label{sec:results:lco}
\input{tables/overall_nc_ctrp}
Table~\ref{tbl:overall_nc_ctrp} shows that,
overall, \SelNet consistently outperforms all other methods in most metrics.
Specifically, \SelNet achieved the best $\ah@K$ scores
with impressive results 
of 2.7142, 4.3119, 7.7577, 12.7728, and 18.3331 for $K =$ 3, 5, 10, 20, and 40, respectively.
Following \SelNet, the other listwise ranking method,
\ListNet achieved the second-best performance in terms of $\ah@K$.
%
%
Overall, \SelNet achieved better {\ah}@10, {\ah}@20 and {\ah}@40 over \ListNet,
%
whereas both methods achieved competitive hit rates up to $K \le 5$.
This suggests that \SelNet is particularly effective in pushing almost all sensitive drugs to the top
while \ListNet was able to push only a few most sensitive drugs.
This is further reflected in the consistent improvements observed in \ah and \ap.
Compared to \ListNet, \SelNet improved 
{\hit}@10, {\hit}@20, and {\hit}@40 for 
33.9\% (55), 44.6\% (72) and 39.5\% (64) of 162 new cell lines by 
3.8\%, 3.0\% and 1.4\%, respectively.  
%
%
%
Such superior performance of \SelNet over \ListNet can be attributed to the ability of \SelNet
to accurately estimate the probability of drugs being sensitive in each cell line
while \ListNet focuses solely on the most sensitive (i.e., top-ranked) drug while ignoring the other sensitive drugs.
%

Furthermore, \SelNet outperformed the best baseline method,
\cclerank, across all metrics.
Moreover, compared to \cclerank, both \SelNet and \ListNet demonstrated
significantly better or competitive performance in \ah and \ap.
This implies that all our proposed methods can improve the ranking performance over \cclerank
by explicitly leveraging auxiliary information such as gene expression profiles and molecular fingerprints.
Specifically, compared to \cclerank,
the best-performing method, \SelNet, 
demonstrated statistically significant improvement in
{\hit}@20, {\hit}@40 and {\hit}@60
for 50.6\% (82), 53.0\% (86) and 45.7\% (74) of 162 new cell lines by 8.6\%, 6.3\% and 5.0\%, respectively,
while
achieving marginally better hit rates for $K < 20$.
Additionally, \SelNet achieved better {\ap}@20 and {\ap}@40 than \cclerank,
improving {\pr}@20 and {\pr}@40 for 53.7\% (87) and 58.0\% (94) of 162 new cell lines by 1.1\% and 3.0\% on average, respectively.
Such consistent and significant improvement across multiple \ah and \ap metrics
on a large percentage of cell lines provides strong evidence that
\SelNet clearly outperforms the best baseline method \cclerank in drug selection and prioritization.

These results suggest that \SelNet can effectively leverage the drug structure information and the entire ranking structure to
learn richer latent representations while
focusing on learning to select all the sensitive drugs in a cell line.
%
The consistent sub-par performance of \cclerank
compared to \SelNet could be due to the fact that 
\cclerank only focuses on the pairwise relative ordering
without considering the overall ranking structure.
Since there are significantly more insensitive drugs than sensitive ones in each cell line,
such pairwise methods may struggle to preserve the ordering between pairs of sensitive and insensitive drugs,
thereby leading to a sub-optimal selection of all sensitive drugs.
%
Overall, all ranking-based methods outperformed 
the state-of-the-art regression model, \deepcdr, across all metrics.
%
This indicates that learning to estimate the exact drug responses while obtaining a lower overall MSE does not
necessarily guarantee accurate score estimation for the sensitive drugs,
which constitutes only 5\% of all drugs in a cell line. 
This leads to sub-par performance of \deepcdr
in terms of selecting and prioritizing the most sensitive drugs in a cell line.

\subsection{Study of Cell Line embeddings}
\label{sec:results:cell}
We evaluated the quality of cell line embeddings based on their ability to capture the drug response profiles.
In order to quantitatively evaluate this,
we computed the pairwise similarities of cell lines in two different ways:
1) using the radial basis function (RBF) kernel on the
learned cell line embeddings out of the best baseline, \cclerank, and the best method, \SelNet,
denoted as \csimB and \csimM, respectively; and
2) using Spearman rank correlation on the ranked lists of drugs
given their drug response profiles, denoted as \csimS.
Intuitively, \csimB and \csimM are higher for pairs of cell lines that are 
close in their corresponding latent spaces.
Meanwhile, \csimS is higher for pairs of cell lines if they share similar drug response profiles.
%
Note that since every drug may not have recorded responses in each cell line,
\csimS for a pair of cell lines $p$ and $q$ is computed
from drug response data of the shared set of drugs, $\SetDrugs_{p} \cap \SetDrugs_{q}$,
whose responses are recorded for both cell lines.

We hypothesize that the latent space captures the ranking structures across drugs,
implying that the cell lines close in the latent space have similar drug response profiles
(i.e., \csimB and \csimM are well correlated with
\csimS).
%
In order to validate our hypothesis, we calculated the Pearson correlations
between \csimB and \csimS, denoted as $\mathtt{corr_c(B, R)}$,
and between \csimM and \csimS, denoted as $\mathtt{corr_c(M, R)}$.
%
We observed that the pairwise cell line similarities induced by
their drug response profiles are better correlated
with the similarities induced in the latent space learned by \SelNet compared to \cclerank
(Pearson correlations $\mathtt{corr_c(M, R)}$ vs. $\mathtt{corr_c(B, R)}$: 0.162 vs. 0.151). 
%
%
This suggests that \SelNet learns informative cell line embeddings that
can capture the overall ranking structure more effectively than \cclerank.
Intuitively, \SelNet may benefit from the fact that it uses the gene expression profile
to learn cell line embeddings;
and because cell lines with similar gene expression profiles typically demonstrate similar drug response profiles.
%
Although \cclerank uses a weighted regularizer to constrain cell line embeddings
based on their genomic similarity
it may not fully capture the complex relationships between the gene expressions of two cell lines.
Explicitly learning embeddings from gene expressions 
allows \SelNet to extract more nuanced task-relevant relationships and a desired notion of similarity between cell lines.
%

\begin{figure}[h]
    \centering
     \includegraphics[width=0.3\textwidth]{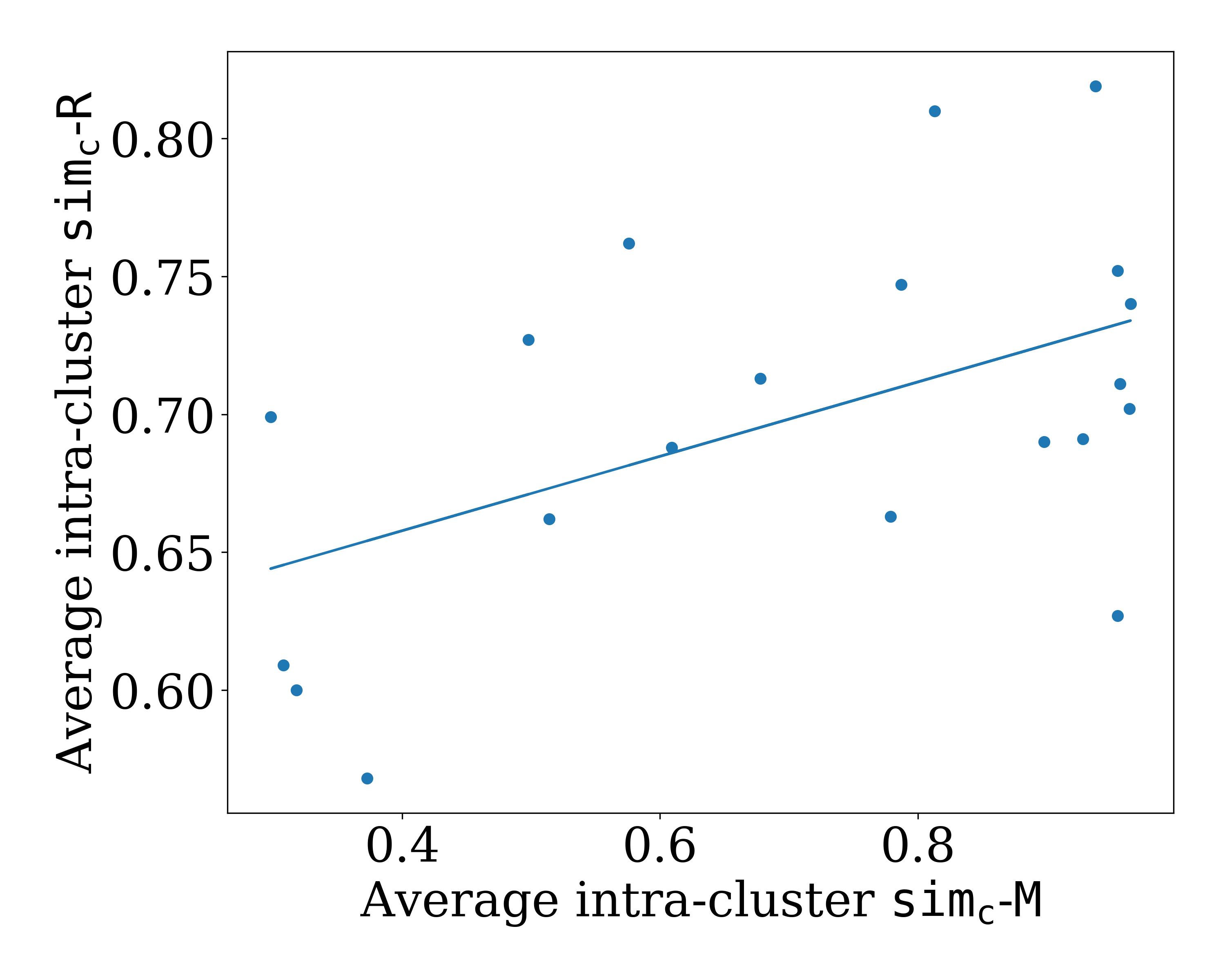}
     \vspace{-5pt}
      \caption{Scatter plot of intra-cluster similarities,
     computed as the average of \csimS and \csimM within each cluster.
      }
      \label{fig:ctrp:cell_isim}
\end{figure}
We further evaluated the quality of their latent spaces in more detail
with respect to clustering compactness and different cancer types.
We applied a 20-way clustering using CLUTO\cite{cluto} 
on the embeddings.
Figure~\ref{fig:ctrp:cell_isim} presents the intra-cluster similarities, i.e., \csimS vs. \csimM
averaged across cell lines in each cluster.
%
We observed that compact clusters in the latent space
contained cell lines with more similar drug-ranking structures.
This is evident from the fitted line with a positive slope as shown in Figure~\ref{fig:ctrp:cell_isim}.
This further supports our hypothesis that the cell line latent space learned by \SelNet effectively 
captures the drug response profiles.
Moreover, our clustering analysis can uncover unobvious or previously unknown similarities among 
cell lines from different cancer types, which may not be apparent from the observed drug response data.

\begin{figure}[h!]
\captionsetup[subfigure]{justification=centering}
    \centering
    \begin{subfigure}[t]{0.275\textwidth}
        \includegraphics[width=\textwidth]{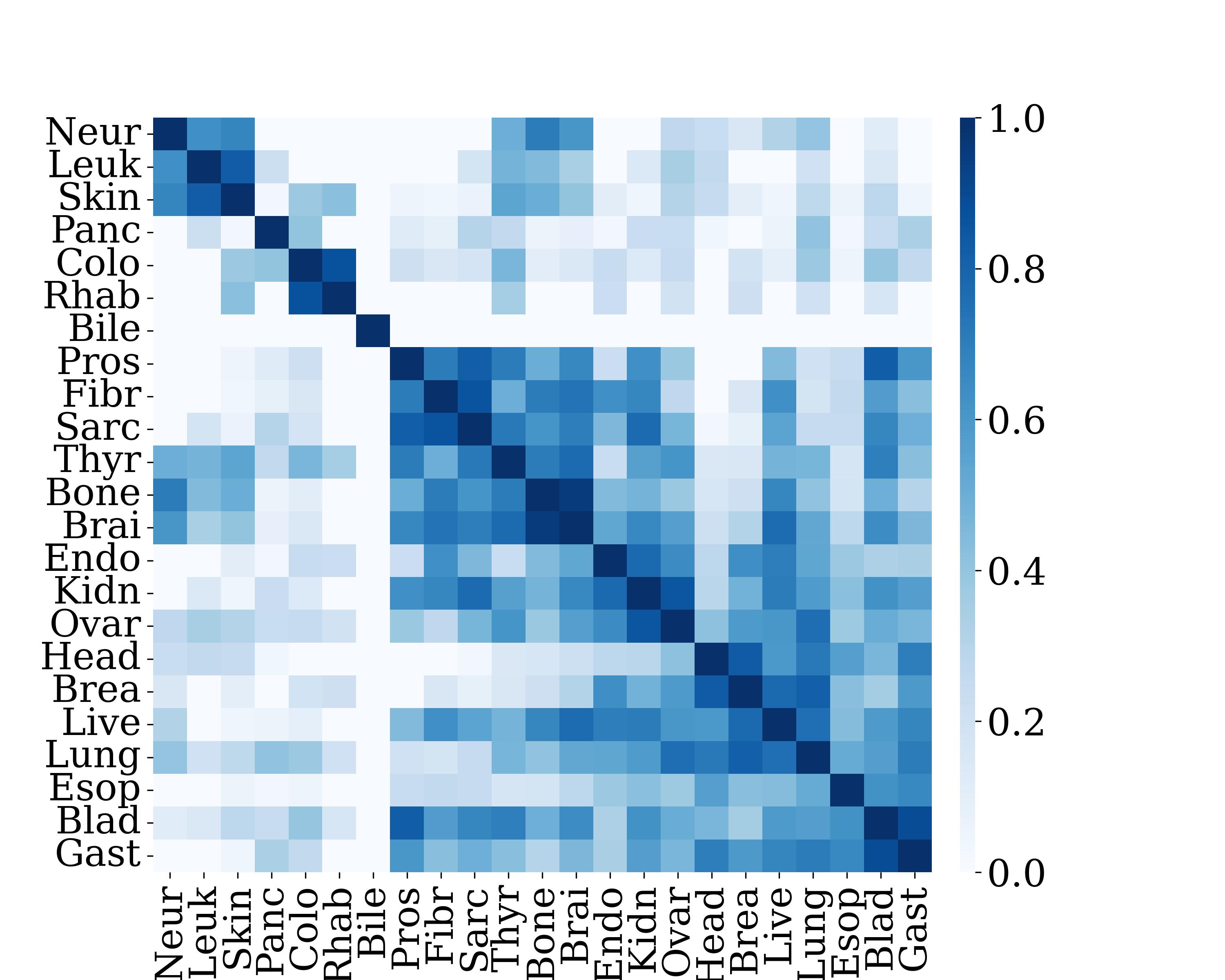}
        \caption{Based on the distribution \\of cell lines
        across clusters.}
    \end{subfigure}%
    ~\hspace{-20pt}
    \begin{subfigure}[t]{0.275\textwidth}
        \includegraphics[width=\textwidth]{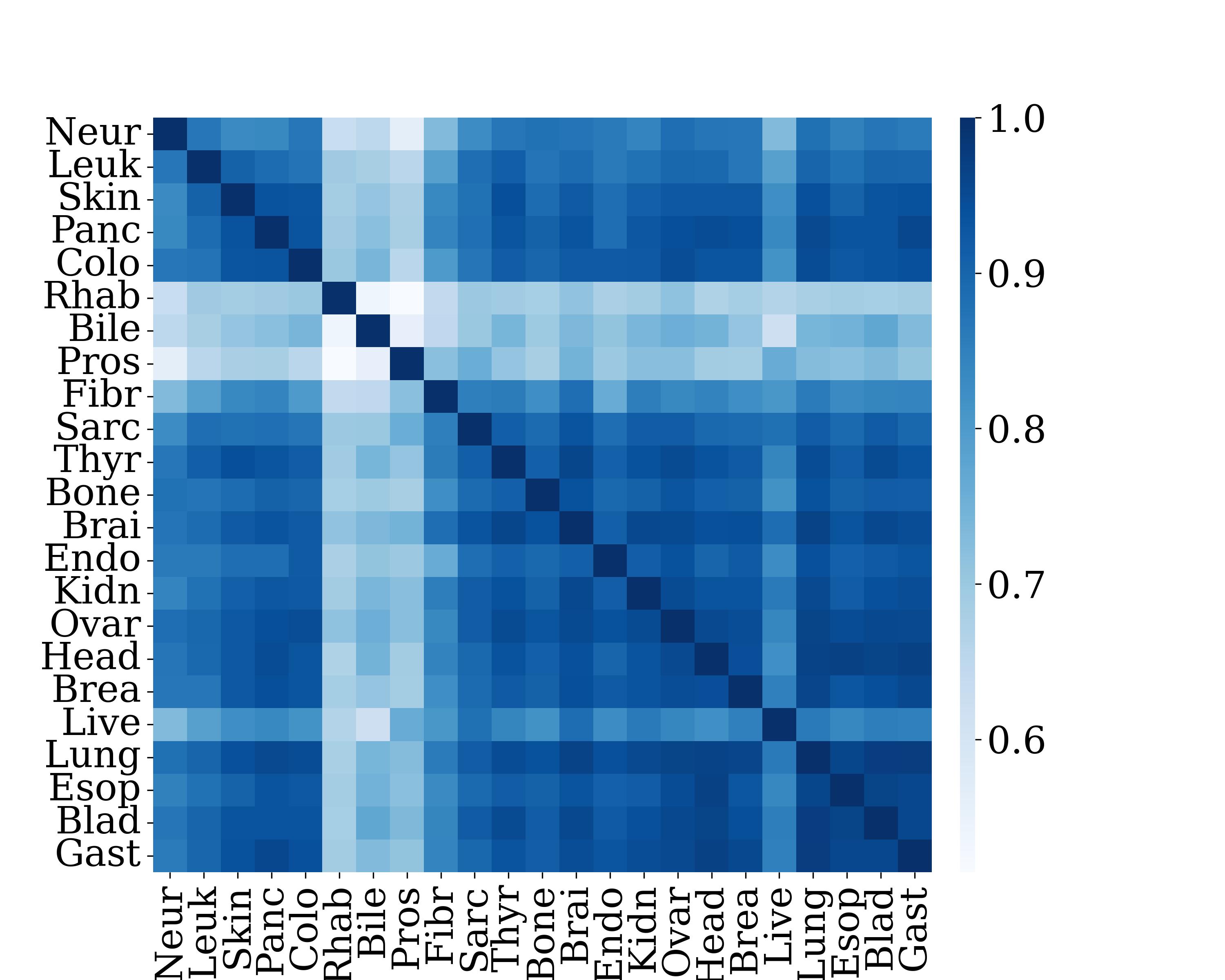}
        \caption{Based on average \csimS \\among cell lines}
    \end{subfigure}
    \vspace{-5pt}
    \caption{Comparison of two pairwise similarity matrices across different cancer types.}
    \label{fig:ctrp:cell_cluster}
\end{figure}

Figure~\ref{fig:ctrp:cell_cluster}a presents the average pairwise similarities
with respect to clustering distribution
of cell lines grouped by different cancer types.
The cell line similarities in this matrix are computed using the Jaccard coefficient on
the normalized distribution of cell lines over the top-10 compact clusters. 
Intuitively, the color in each cell in Figure~\ref{fig:ctrp:cell_cluster}a indicates the 
degree of clustering overlap between cell lines from two different cancer types.
In other words, if cell lines from two different cancer types are
clustered together or distributed identically over multiple clusters, 
they will be more similar
and will have darker shades in the respective cell in this figure.
%
For example, the cell lines from kidney and ovary cancer types
are often clustered together, such is the case for bladder and gastric cancer types.
Figure~\ref{fig:ctrp:cell_cluster}b presents the average pairwise \csimS similarities
among cell lines from different cancer types.
Specifically, if cell lines from two different cancer types share similar
drug ranking structures (i.e., high \csimS) on average,
the corresponding pair of cancer types in this figure will have a darker shade.
%

Overall, we observed a moderate correlation between the cluster overlap-based similarities (Figure~\ref{fig:ctrp:cell_cluster}a)
and the drug ranking structure-based similarities (Figure~\ref{fig:ctrp:cell_cluster}b)
with a Pearson correlation of 0.493 ($p$-value = 1e-33). 
Additionally, Figure~\ref{fig:ctrp:cell_cluster}a can provide clinically significant and valuable insights
while uncovering similarities between cell lines of
different cancer types even though their drug-ranking structures or drug response profiles do not
exhibit significant similarities.
%
For instance, the liver cancer cell lines tend to be clustered
with cell lines of different cancer types such as bone, brain, breast and lung cancers (Figure~\ref{fig:ctrp:cell_cluster}a) even though
their drug ranking structures are apparently different
(Figure~\ref{fig:ctrp:cell_cluster}b).
As a matter of fact, several studies\cite{Ananthakrishnan2006, Riihimki2018} in the medical literature
provide evidence of secondary liver cancers (i.e., metastatic liver tumors)
spreading from primary tumors of breast and lung origins.
Moreover, the most common mutations causing liver cancer
(namely, TP53, CTNNB1, AXIN1, ARID1A, CDKN2A and CCND1 genes)\cite{Khemlina2017} 
are commonly associated with multiple cancers.
%
Similar observations can be made from the figure
for prostate, bladder, sarcoma, and thyroid cancers,
which also tend to be co-occurring according to reports in the literature~\cite{Tomaszewski2014,Davis2015,Fan2017}.
%
%

We further validate that the cell line latent space in \SelNet
is capable of grouping cell lines based on their cancer types, and in fact, does so better
than that in \cclerank.
%
%
In order to validate this, we calculated $k$-nearest-neighbor accuracy of a cell line $c$ in the latent space,
denoted as $\acc(c)$, 
as follows:
\begin{equation}
\acc(c) = \frac{1}{k}\sum_{\scriptsize{c' \in \mathtt{kNN}(c,k)}}
\mathbb{I}[\mathtt{Cancer}(c') = \mathtt{Cancer}(c)],
\end{equation}
where $\mathtt{kNN}(c,k)$ returns $k$-nearest neighbors of a cell line $c$ in the 
latent space, $\mathbb{I}$ is the indicator function, 
and $\mathtt{Cancer}(c)$ returns the cancer type of cell line $c$.
Specifically, $\acc(c)$ is the expected fraction of $k$ nearest neighboring cell lines
that share the same cancer type as the cell line $c$.
%
We observed that the average $\acc$ over all unseen cell lines
were higher in \SelNet compared to \cclerank
(\SelNet vs. \cclerank:  0.364 vs. 0.162, 0.181 vs. 0.110, 0.126 vs 0.082) for $k=1,3,5$, respectively. 
This suggests that the latent space in \SelNet is better clustered
with respect to cancer types, even though the cancer type information is never
fed to the model.
This is likely due to the fact that \SelNet
incorporates the gene expression profile during pretraining, 
and typically cell lines from the same cancer type tend to share similar gene expression profiles.
In summary, not only the latent space in \SelNet clusters cell lines based on their drug ranking structures,
it also maps cell lines from the same cancer types (i.e., of the same origin) to close proximity.
These properties of the latent space can have potential clinical applications,
such as determining cancer types for cell lines with unknown origin,
and matching such cell lines with those having known cancer types for additional wet-lab experiments.
%

\subsection{Study of Drug embeddings}
\label{sec:results:drug}
We evaluated the quality of drug embeddings based on the extent to how well the latent space 
captures the sensitivity profiles of drugs across cell lines.
The sensitivity profile of a drug was defined as a binary embedding,
with a value of 1 indicating that the drug is sensitive in a cell line, and 0 if insensitive. 
%
%
To quantitatively evaluate the quality of the latent space,
we calculated the pairwise similarities of drugs in two ways:
1) using the RBF kernel on the learned drug embeddings out of the best baseline method, \cclerank,
and the best method, \SelNet, denoted as \dsimB and \dsimM, respectively; and,
2) using the Jaccard coefficient on the corresponding sensitivity profiles of drugs across cell lines, denoted as \dsimS.
Clearly, \dsimB and \dsimM are higher for drug pairs that are 
close in their corresponding latent spaces;
\dsimS is higher for drug pairs if they share similar sensitivity profiles across
many cell lines.
%
It is important to note that not all drugs have recorded responses in each cell line.
Thus, \dsimS for a pair of drugs $p$ and $q$ was calculated
from the sensitivity profiles of the shared set of cell lines
for which the responses were recorded for both drugs.
\begin{figure}[h!]
    \centering
     \includegraphics[width=0.3\textwidth]{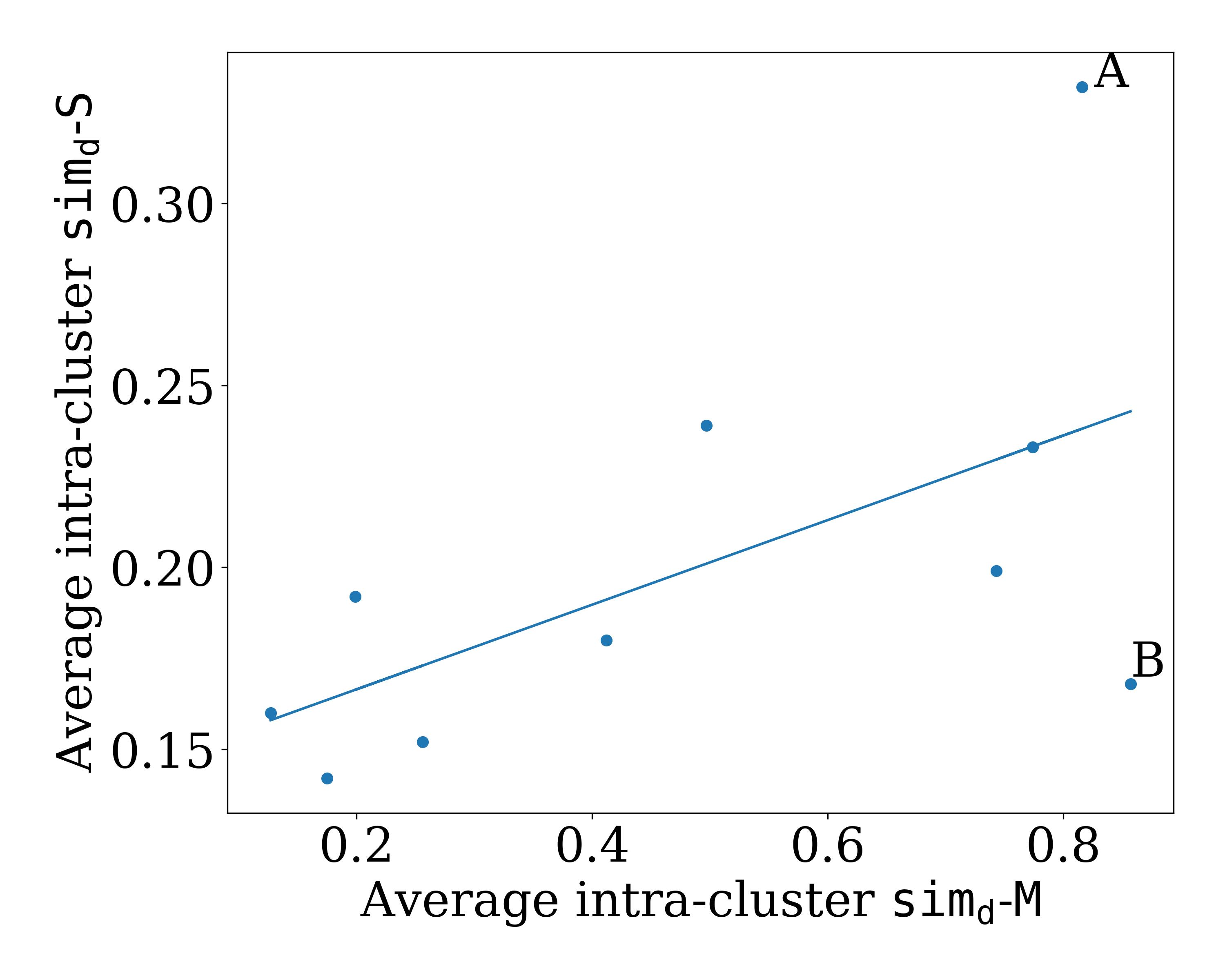}
     \vspace{-5pt}
      \caption{Scatter plot of intra-cluster similarities,
     computed as the average of \dsimS and \dsimM within each cluster.
      }
      \label{fig:ctrp:drug_isim}
\end{figure}
We hypothesize that the learned latent space locally captures the sensitivity profiles of drugs.
In other words, drugs that are close in the latent space have similar sensitivity profiles,
leading to a correlation between the similarities in the latent space (i.e., \dsimB and \dsimM)
and those computed from the sensitivity profiles (i.e., \dsimS).
To test this hypothesis, we computed the Pearson correlations
among the three similarities as follows:
1) correlation between \dsimB and \dsimS, denoted as $\mathtt{corr_d(B, S)}$; and,
2) correlation between \dsimM and \dsimS, denoted as $\mathtt{corr_d(M, S)}$.
%
We observed that the pairwise drug similarities induced by cell sensitivity profiles 
are better correlated to the pairwise similarities induced in the latent space
learned by \SelNet compared to \cclerank 
(Pearson correlations $\mathtt{corr_d(M, S)}$ and $\mathtt{corr_d(B, S)}$: 0.906 vs. 0.352). 
This suggests that \SelNet can learn effective drug embeddings
that can better capture the sensitivity profiles compared to \cclerank.
This may be due to the fact that
\SelNet leverages molecular fingerprints, unlike \cclerank,
to learn drug embeddings that can encode structural information;
and it is well known that structurally similar drugs tend to exhibit similar sensitivities.
%

%
%
%

\begin{figure}[h!]
\captionsetup[subfigure]{justification=centering}
    \centering
    \begin{subfigure}[t]{0.27\textwidth}
        \centering
        \includegraphics[width=\textwidth]{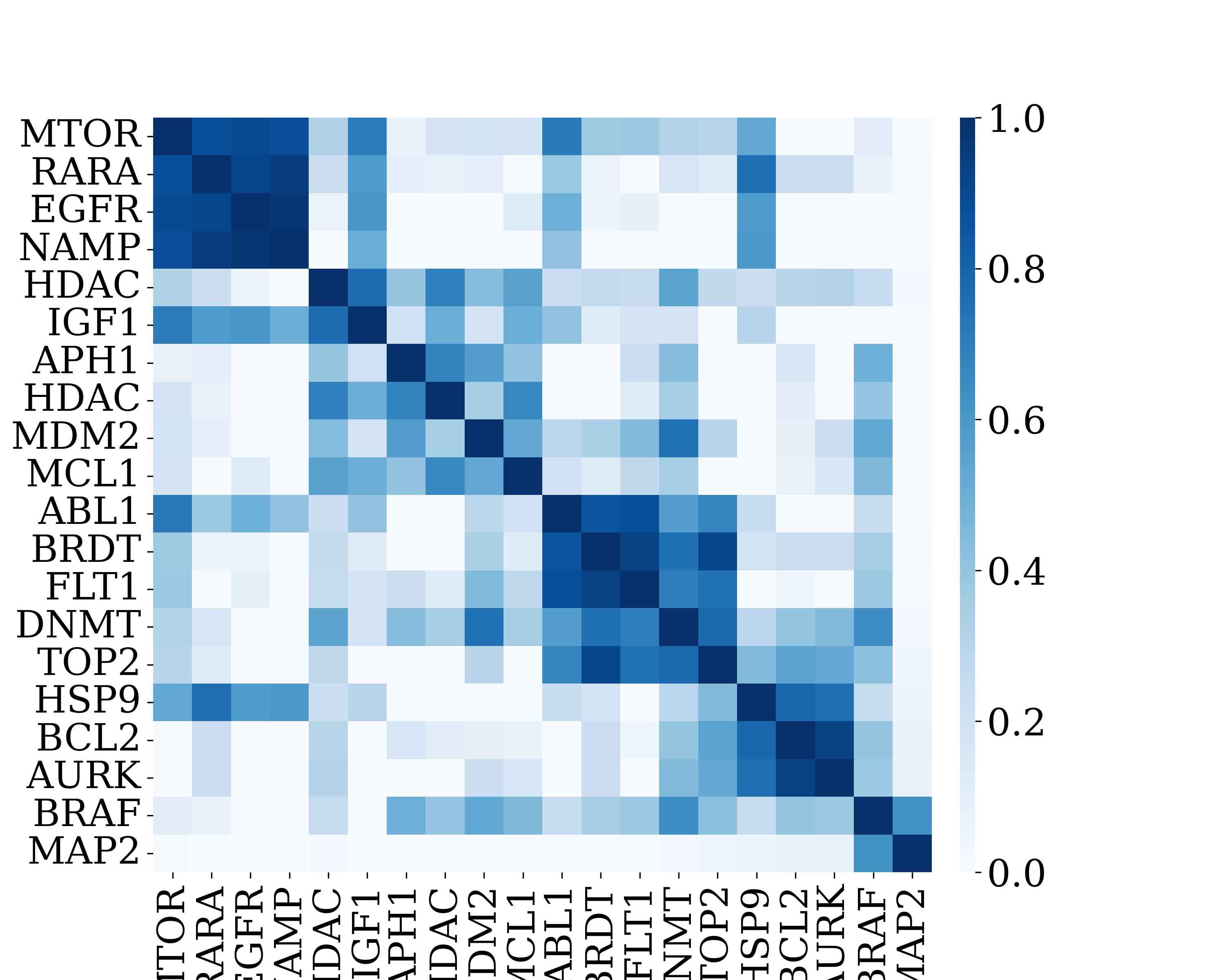}
        \caption{Based on the distribution \\of drugs
        across clusters.}
    \end{subfigure}%
   ~\hspace{-20pt} 
    \begin{subfigure}[t]{0.27\textwidth}
        \centering
        \includegraphics[width=\textwidth]{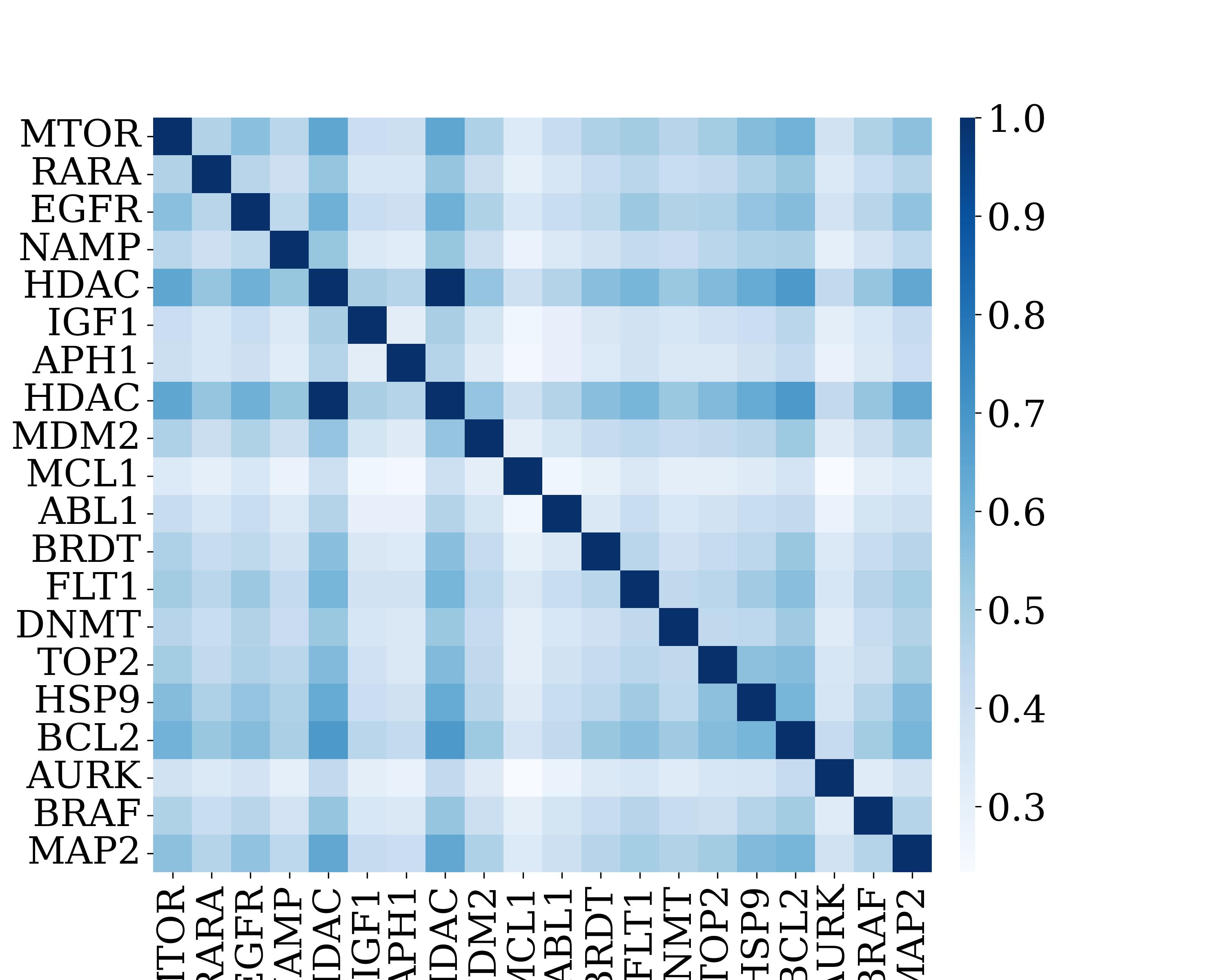}
        \caption{Based on average \dsimS \\among drugs}
    \end{subfigure}
    \vspace{-5pt}
    \caption{Comparison of two pairwise similarity matrices across different MoAs.}
    \label{fig:ctrp:drug_cluster}
\end{figure}
Furthermore, we evaluated the quality of drug embeddings out of \SelNet  
via clustering.
We applied a 10-way clustering (using CLUTO) on the drug embeddings.
Figure~\ref{fig:ctrp:drug_isim} presents the intra-cluster similarities, \dsimS vs. \dsimM
averaged across all drugs in each cluster.
%
We observed that the compact clusters in the latent space
contained drugs with similar sensitivity profiles.
This further supports our previous hypothesis that the latent space for drugs effectively 
captures the sensitivity profiles.

Furthermore, we studied the drug clusters in more detail and identified some
qualities of the latent space with respect to uncovering the mechanism of action (MoA) of drugs. 
Figure~\ref{fig:ctrp:drug_cluster}a presents the average pairwise similarities
among drugs grouped by different MoAs, where
the similarities are computed using the Jaccard coefficient on the
normalized distribution of drugs across clusters (Figure~\ref{fig:ctrp:drug_isim}).
In other words, if drugs with different MoAs are clustered together or
co-occurs over multiple clusters, they are considered similar
and have darker shades in the respective cells in this figure.
Figure~\ref{fig:ctrp:drug_cluster}b presents the average pairwise \dsimS similarities
among drugs with different MoAs. 
Notably, we find that certain MoAs, such as MTOR, RARA, EGFR, and NAMP,
exhibit similarities in their clustering patterns, suggesting potential shared characteristics or pathways.
Similarly, we observe similarities among ABL1, BRDT, and FLT1, as well as BCL2 and AURK.
These findings might indicate potential commonalities among drugs with different MoAs, 
even when their sensitivity profiles may not be similar (Figure \ref{fig:ctrp:drug_cluster}b). 
%
%
\begin{figure}[h!]
    \centering
    \begin{subfigure}[h]{0.23\textwidth}
        \centering
        \includegraphics[width=\textwidth]{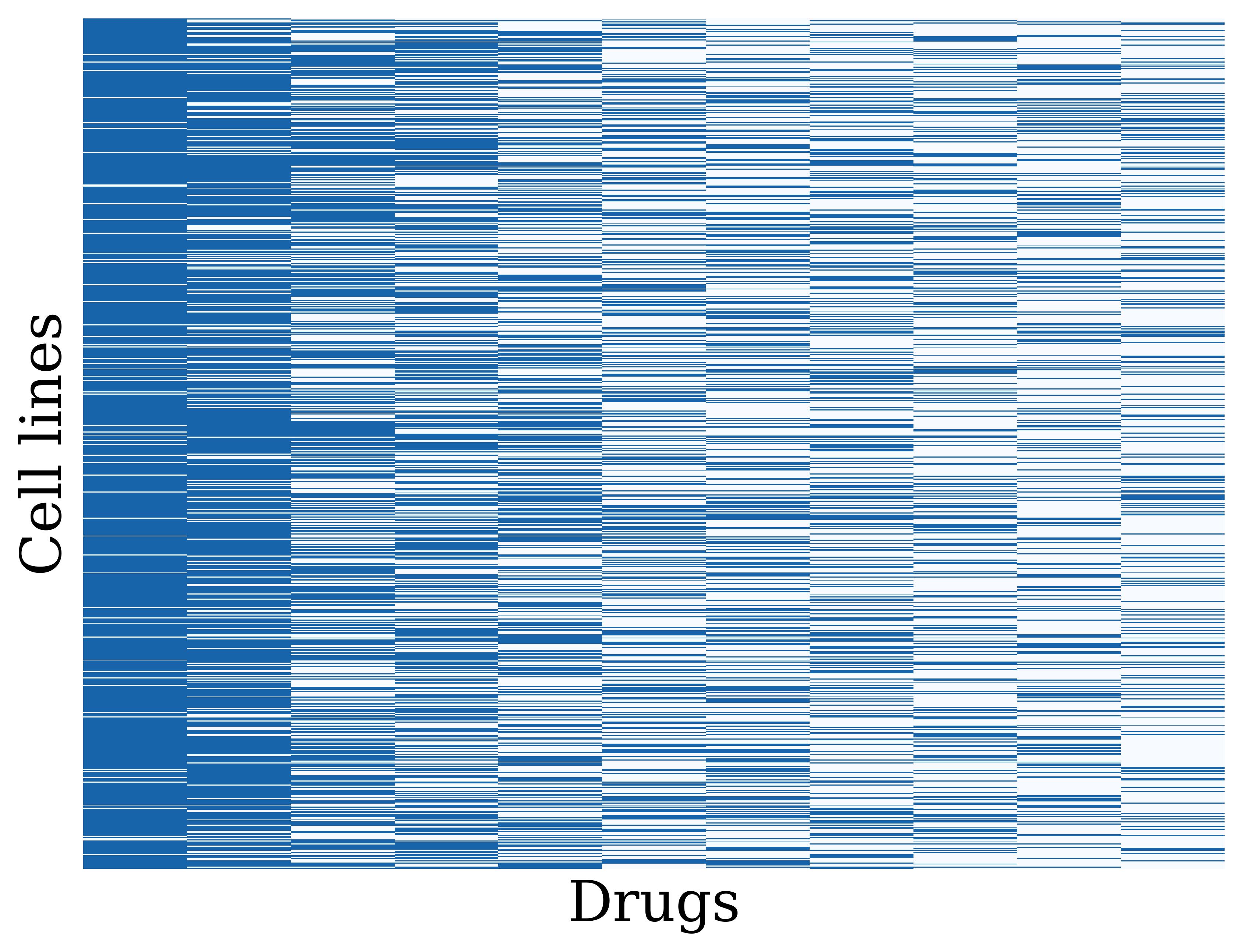}
        \caption{Cluster A}
    \end{subfigure}%
    ~ 
    \begin{subfigure}[h]{0.23\textwidth}
        \centering
        \includegraphics[width=\textwidth]{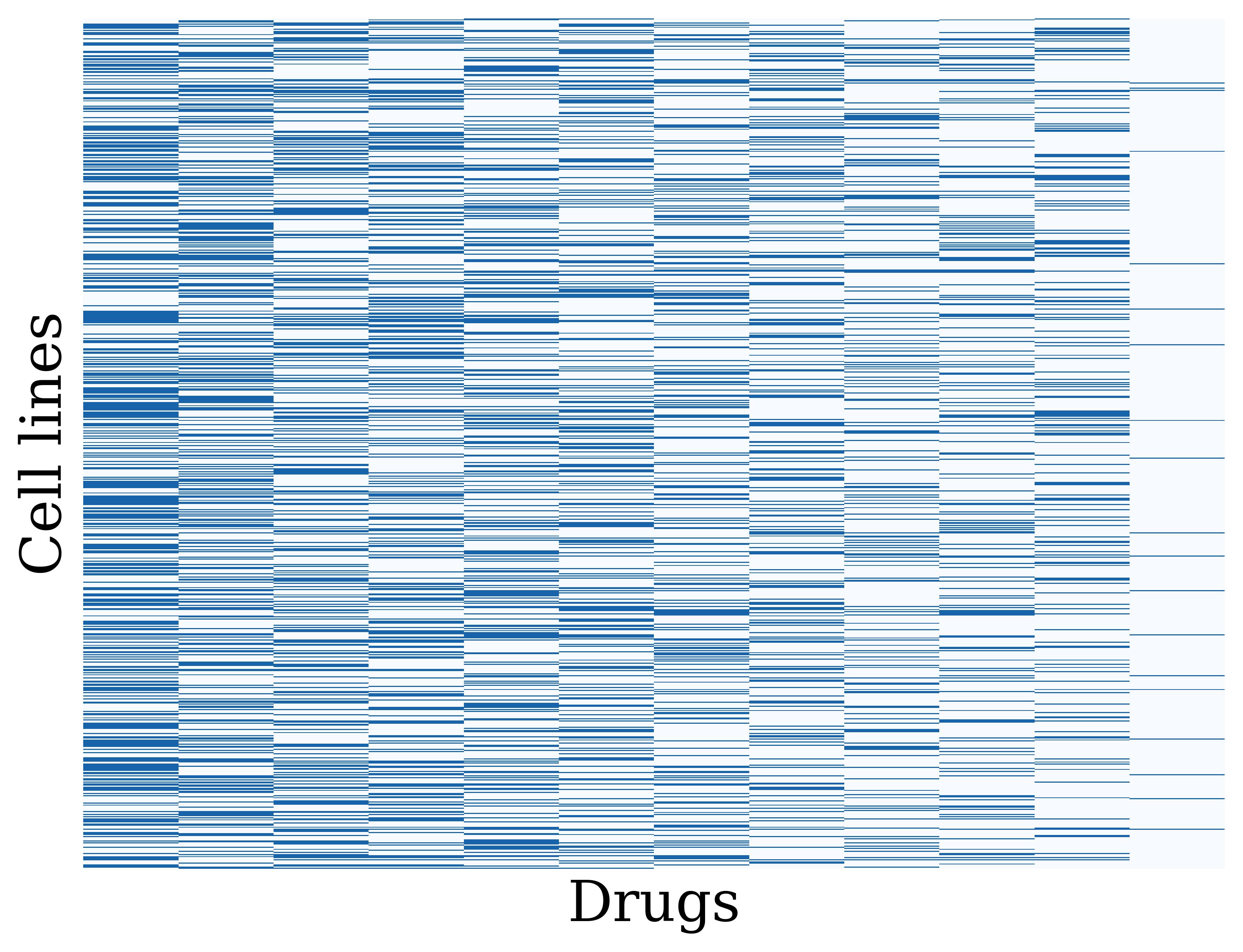}
        \caption{Cluster B}
    \end{subfigure}
    \vspace{-5pt}
     \caption{Comparison of sensitivity profiles for all drugs in clusters A and B in Figure~\ref{fig:ctrp:drug_isim}.
     Darker color indicates that the drug is sensitive in corresponding cell lines.}
    \label{fig:ctrp:drug_clusters}
\end{figure}

%
We further examined clusters A and B depicted in Figure~\ref{fig:ctrp:drug_isim}
to gain deeper insights into their characteristics.
Despite both clusters being compact,
they exhibit distinct characteristics in terms of their \dsimS similarities.
%
%
Figure~\ref{fig:ctrp:drug_clusters} presents the sensitivity profiles of
all the drugs in each cluster.
Clearly, compared to cluster B,
most drugs in cluster A 
share multiple cell lines in which they are sensitive, thus resulting in higher \dsimS for cluster A than for cluster B.
%
Specifically, from our preliminary fact-checking, we found that
many drugs in cluster A (left-most drugs in Figure~\ref{fig:ctrp:drug_clusters}a)
share some common pathways such as EGFR, mTOR.
This suggests that these drugs possess broader effectiveness across multiple cancer types. 
%
%
%
%
%
In summary, our analysis reveals that the latent drug space learned by \SelNet
captures similarities in sensitivity profiles, molecular structures, and pathway mechanisms.
These findings highlight the potential for exploring synergistic effects among drugs with different MoAs,
and developing novel therapeutic strategies.

\section{Conclusion}
\label{sec:conclusion}

In this work, we developed two listwise neural ranking methods 
to select anti-cancer drugs out of all known drugs for new cell lines.
%
Our experiments suggest that our listwise ranking method, \SelNet,
can select all the sensitive drugs instead of the few top-most sensitive drugs.
Moreover, our experimental comparison with strong ranking and regression baselines demonstrated the
efficacy of
formulating drug selection as a LeToR problem. 
%
Notably, our method, \SelNet demonstrated significant
improvements over the baseline \cclerank in average hit rates
across a large proportion of cell lines.
Additionally, by leveraging deep networks and pretraining techniques,
our methods can learn informative embeddings.
%
Our analyses of such learned embeddings revealed commonalities
among cell lines and among drugs from different cancer types and MoAs, respectively.
%
Overall, our work represents a step forward in the development of robust and effective methods for precision anti-cancer drug selection.
Future work may explore on leveraging 3D molecular structures, multiple modalities or
pretrained chemical foundational models to further enhance ranking performance.
%

\section{Code Availibility}
\label{sec:code}
The processed data and code are publicly available at \url{https://github.com/ninglab/DrugRanker}.
All the required softwares to execute the code are freely available.

\bibliographystyle{ACM-Reference-Format}
\bibliography{paper}

\appendix
\section{Reproducibility}
As the encoder \GeneE, we used a fully-connected neural network
consisting of 2 hidden layers with 4,096 and 1,024 units, respectively; 
each hidden layer is followed by a ReLU activation.
The last hidden layer is followed by an output layer with 128 units which outputs the cell line latent embedding
(from preliminary experiments, the best results were obtained with 128 units).
We implemented \DrugE as a fully-connected neural network
with one hidden layer of 128 hidden units followed by ReLU non-linearity,
and the output layer of $M$ units, where $M \in \{25,50,100,200\}$.
The hyperparameters for all the baselines and proposed methods were systematically fine-tuned
via a random grid search.
%
The learning rate and learning rate annealing scheme for \cclerank were in accordance with its original implementation.
%
Other methods were trained using ADAM optimization with an initial learning rate of 0.001.
All the baselines were trained for 100 epochs or until the convergence of their respective training objectives.
In contrast, the listwise ranking models were trained for 300 epochs.
Each experiment was conducted on a computing node equipped with NVIDIA Volta V100 GPUs and Dual Intel Xeon 8268s processors.
The hyperparameters for all the methods are reported in Table~\ref{tbl:hyperparam} to facilitate the reproducibility of results.

\input{tables/hyperparameters}

\end{document}

%% file: define.tex
\newcommand{\cclerank}{\mbox{$\mathop{\mathtt{pLETORg}}\limits$}\xspace}
\newcommand{\deepcdr}{\mbox{$\mathop{\mathtt{DeepCDR}}\limits$}\xspace}

\newcommand{\ListNet}{\mbox{$\mathop{\mathtt{List\text{-}One}}\limits$}\xspace}
\newcommand{\SelNet}{\mbox{$\mathop{\mathtt{List\text{-}All}}\limits$}\xspace}

\newcommand{\GeneAE}{\mbox{$\mathop{\mathtt{GeneAE}}\limits$}\xspace}
\newcommand{\GeneE}{\mbox{$\mathop{\mathtt{GeneE}}\limits$}\xspace}
\newcommand{\GeneD}{\mbox{$\mathop{\mathtt{GeneD}}\limits$}\xspace}
\newcommand{\DrugE}{\mbox{$\mathop{\mathtt{DrugE}}\limits$}\xspace}

\newcommand{\SetCells}{\mbox{$\mathcal{C}$}\xspace}
\newcommand{\SetDrugs}{\mbox{$\mathcal{D}$}\xspace}
\newcommand{\CellEmb}{\mbox{$\mathbf{u}$}\xspace}
\newcommand{\DrugEmb}{\mbox{$\mathbf{v}$}\xspace}

\newcommand{\ci}{\mbox{$\mathop{\mathtt{CI}}\limits$}\xspace}
\newcommand{\sci}{\mbox{$\mathop{\mathtt{sCI}}\limits$}\xspace}
\newcommand{\pr}{\mbox{$\mathop{\mathtt{P}}\limits$}\xspace}
\newcommand{\ap}{\mbox{$\mathop{\mathtt{AP}}\limits$}\xspace}
\newcommand{\hit}{\mbox{$\mathop{\mathtt{H}}\limits$}\xspace}
\newcommand{\ah}{\mbox{$\mathop{\mathtt{AH}}\limits$}\xspace}

\newcommand{\csimB}{\mbox{$\mathop{\mathtt{sim_c\text{-}{B}}}\limits$}\xspace}
\newcommand{\csimM}{\mbox{$\mathop{\mathtt{sim_c\text{-}{M}}}\limits$}\xspace}
\newcommand{\csimS}{\mbox{$\mathop{\mathtt{sim_c\text{-}{R}}}\limits$}\xspace}
\newcommand{\dsimB}{\mbox{$\mathop{\mathtt{sim_d\text{-}{B}}}\limits$}\xspace}
\newcommand{\dsimM}{\mbox{$\mathop{\mathtt{sim_d\text{-}{M}}}\limits$}\xspace}
\newcommand{\dsimS}{\mbox{$\mathop{\mathtt{sim_d\text{-}{S}}}\limits$}\xspace}

\newcommand{\acc}{\mbox{$\mathop{\mathtt{acc_{\scriptsize{kNN}}}}\limits$}\xspace}

%% file: tables/notation.tex
\begin{table}[h]
\centering
\caption{Notations and Definitions}
\label{tbl:notations}
\vspace{-10pt}
\begin{threeparttable}
	 \begin{tabular}{
	 	@{\hspace{2pt}}l@{\hspace{2pt}}
	 	@{\hspace{2pt}}p{6.5cm}@{\hspace{2pt}}
		}
		\toprule
		Notation & Definition \\         
		\midrule
         \SetCells & Set of cell lines \\
         \SetDrugs & Set of drugs \\
         $d_i$ & Drug $i$ \\
         $\SetDrugs^{+}_c$ / $\SetDrugs^{-}_c$ & Set of sensitive/insensitive drugs for a cell line $c$ \\
         $\CellEmb_c$ / $\DrugEmb_d$ & embedding for cell line $c$/drug $d$ \\
         \bottomrule
   
         \end{tabular}
\end{threeparttable}
\vspace{-5pt}

\end{table}

%% file: tables/data.tex
\begin{table}[h]
\centering
\caption{Dataset Overview}
\label{tbl:data}
\vspace{-10pt}
  \begin{threeparttable}
      \begin{tabular}{
          @{\hspace{8pt}}l@{\hspace{8pt}}
          @{\hspace{8pt}}r@{\hspace{8pt}}
          @{\hspace{8pt}}r@{\hspace{8pt}}
          @{\hspace{8pt}}r@{\hspace{8pt}}
          @{\hspace{8pt}}r@{\hspace{8pt}}
          @{\hspace{8pt}}r@{\hspace{8pt}}
          @{\hspace{8pt}}r@{\hspace{8pt}}
          }
          \toprule
          Dataset & |\SetCells| & |\SetDrugs| & \#AUCs & \#d/{\SetCells} & c/{\SetDrugs} & m\% \\
          \midrule
          CTRP & 809 & 545 & 357,544 & 442 & 656 & 18.9 \\ 
          \bottomrule
          
          \end{tabular}
 \begin{tablenotes}
\footnotesize
        \setlength\labelsep{0pt}
		\item In this table,  the columns $|\SetCells|$, $|\SetDrugs|$, and \#AUCs denote the number of 
		unique cell lines, drugs and cell-line drug response pairs, respectively. 
		The columns \#d/{\SetCells}, c/{\SetDrugs} and m\% denote the average number of drugs per cell line,
		average number of cell lines per drug, and the missing percentage of responses, respectively.
		\par

\end{tablenotes}

\end{threeparttable}

\end{table}

%% file: tables/overall_nc_ctrp.tex
\begin{table*}[h]
\centering
\caption{Overall Comparison for CTRP in LCO validation}
\label{tbl:overall_nc_ctrp}
\vspace{-10pt}
  \begin{threeparttable}
      \begin{tabular}{
          @{\hspace{2pt}}l@{\hspace{1pt}}
          @{\hspace{4pt}}r@{\hspace{4pt}}
          @{\hspace{4pt}}r@{\hspace{4pt}}
          @{\hspace{4pt}}r@{\hspace{4pt}}
          @{\hspace{4pt}}r@{\hspace{4pt}}
          @{\hspace{4pt}}r@{\hspace{4pt}}
          @{\hspace{4pt}}r@{\hspace{4pt}}
          @{\hspace{4pt}}r@{\hspace{4pt}}
          @{\hspace{4pt}}r@{\hspace{4pt}}
          @{\hspace{4pt}}r@{\hspace{4pt}}
	  @{\hspace{4pt}}r@{\hspace{4pt}}
	  @{\hspace{4pt}}r@{\hspace{4pt}}
          @{\hspace{4pt}}r@{\hspace{4pt}}
          @{\hspace{4pt}}r@{\hspace{4pt}}
          @{\hspace{4pt}}r@{\hspace{4pt}}
          }
          \toprule
	  model  & AP@1 &  AP@3   & AH@3   & AP@5   & AH@5   & AP@10  & AH@10  
	  & AP@20  & AH@20   & AP@40  & AH@40   & AP@60  & AH@60  \\ 
	  \midrule
	\cclerank & 0.9308\rlap{*} & 0.9586 & 2.6853 & 0.9391 & 4.2454 & 0.8962 & 7.3525 
	& 0.8178 & 11.9248 & 0.7286 & 17.1087 & 0.6877 & 19.5646 \\
	\cclerank & 0.9306 & 0.9598\rlap{*} & 2.6944 & 0.9402 & 4.2756 & 0.9018 & 7.4802 
	& 0.8255 & 12.0889 & 0.7361 & 17.3179 & 0.7011 & 19.3712 \\
	\cclerank & 0.9306 & 0.9596 & 2.6968\rlap{*} & 0.9402 & 4.2731 & 0.9018 & 7.4716 
	& 0.8250 & 12.0936 & 0.7351 & 17.3506\rlap{*} & 0.7002 & 19.4150 \\
	\cclerank & 0.9306 & 0.9593 & 2.6951 & 0.9410\rlap{*} & 4.2582 & 0.8999 & 7.3960 
	& 0.8222 & 12.0270 & 0.7328 & 17.2086 & 0.6951 & 19.4621 \\
	\cclerank & 0.9306 & 0.9597 & 2.6943 & 0.9402 & 4.2829\rlap{*} & 0.9023 & 7.4804\rlap{*} 
	& 0.8257\rlap{*} & 12.0962\rlap{*} & 0.7364\rlap{*} & 17.3279 & 0.7023\rlap{*} & 19.3091 \\
	\cclerank & 0.9306 & 0.9595 & 2.6878 & 0.9395 & 4.2791 & {0.9026}\rlap{*} & 7.4467 
	& 0.8246 & 12.0456 & 0.7353 & 17.2695 & 0.6995 & 19.3624 \\
	\cclerank & 0.9306 & 0.9587 & 2.6748 & 0.9376 & 4.2236 & 0.8988 & 7.3600 
	& 0.8192 & 12.0065 & 0.7304 & 17.2451 & 0.6893 & 19.7447\rlap{*} \\
	\hline
	  \deepcdr & 0.9260\rlap{*} & 0.9296\rlap{*} & 2.4716 & 0.9015 & 4.0904 & 0.8646 & 7.4440\rlap{*} 
	& 0.8025 & 11.5779 & 0.7155 & 16.7472\rlap{*} & 0.6736 & 19.3054\rlap{*} \\
	\deepcdr & 0.9129 & 0.9271 & 2.5151\rlap{*} & 0.9035 & 4.1096\rlap{*} & 0.8665\rlap{*} & 7.4384 
	& 0.8036\rlap{*} & 11.7047\rlap{*} & 0.7211\rlap{*} & 16.7447 & 0.6791\rlap{*} & 19.3008 \\
	\deepcdr & 0.9209 & 0.9282 & 2.5087 & 0.9064\rlap{*} & 4.0324 & 0.8599 & 7.3529 
	& 0.7984 & 11.5311 & 0.7118 & 16.5821 & 0.6700 & 19.2082 \\
	\hline
	\ListNet & 0.9478\rlap{*}  & 0.9523\rlap{*}  & 2.5284 & 0.9170 & 3.9832 & 0.8614 & 7.2736 
	& 0.7929 & 12.0092 & 0.7152 & 17.4724 & 0.6773 & 20.0047 \\
	\ListNet & 0.9359 & 0.9499 & 2.6392\rlap{*}  & 0.9278 & 4.1952\rlap{*} & 0.8830 & 7.4661 
	& 0.8128 & 12.3445 & 0.7344 & 17.6828 & 0.6963 & 20.1293 \\
	\ListNet & 0.9423 & 0.9507 & 2.6285 & 0.9293\rlap{*}  & 4.1583 & 0.8833\rlap{*} & 7.4334 
	& 0.8115 & 12.1450 & 0.7304 & 17.5116 & 0.6902 & 20.0350 \\
	\ListNet & 0.9393 & 0.9485 & 2.6315 & 0.9269 & 4.1909 & 0.8828 & 7.5999\rlap{*}  
	& 0.8174\rlap{*}  & 12.4297 & 0.7389\rlap{*}  & 17.9255 & 0.7024\rlap{*}  & 20.2904 \\
	\ListNet & 0.9165 & 0.9371 & 2.5749 & 0.9152 & 4.1013 & 0.8694 & 7.5055 
	& 0.8065 & 12.5335\rlap{*}  & 0.7330 & 17.8804 & 0.6970 & 20.2376 \\
	\ListNet & 0.9030 & 0.9300 & 2.5543 & 0.9106 & 4.0587 & 0.8640 & 7.4101 
	& 0.8009 & 12.4205 & 0.7262 & 17.9959\rlap{*}  & 0.6924 & 20.3382\rlap{*} \\
	\hline
\SelNet & \textbf{0.9480}\rlap{*} & \textbf{0.9610}\rlap{*} & 2.6985 & 0.9420 & 4.2885 & 0.9087 & 7.6208 & 0.8333 & 12.1992 & 0.7429 & 17.4686 & 0.7012 & 20.0028 \\
\SelNet & 0.9295 & 0.9537 & \textbf{2.7142}\rlap{*} & 0.9403 & 4.3000 & 0.9060 & 7.6104 & 0.8327 & 12.2481 & 0.7442 & 17.5094 & 0.7035 & 19.9869  \\
\SelNet & 0.9421 & 0.9597 & 2.6984 & \textbf{0.9442}\rlap{*} & 4.2919 & 0.9094 & 7.6224 & 0.8345 & 12.1688 & 0.7436 & 17.4484 & 0.7020 & 19.9981  \\
\SelNet & 0.9365 & 0.9604 & 2.6934 & 0.9411 & \textbf{4.3119}\rlap{*} & 0.9067 & 7.6108 & 0.8334 & 12.1745 & 0.7429 & 17.4458 & 0.7014 & 19.9993  \\
\SelNet & 0.9395 & 0.9581 & 2.7076 & 0.9414 & 4.3056 & \textbf{0.9102}\rlap{*} & 7.6232 & 0.8339 & 12.1874 & 0.7432 & 17.4571 & 0.7020 & 19.9871  \\
\SelNet & 0.9255 & 0.9452 & 2.6474 & 0.9255 & 4.2547 & 0.8890 & \textbf{7.7577}\rlap{*} & 0.8276 & 12.6216 & 0.7489 & 17.9918 & 0.7116 & 20.3199  \\
\SelNet & 0.9418 & 0.9598 & 2.6903 & 0.9409 & 4.3037 & 0.9067 & 7.6969 & \textbf{0.8361}\rlap{*} & 12.3357 & 0.7478 & 17.6577 & 0.7073 & 20.0894  \\
\SelNet & 0.8586 & 0.9120 & 2.5588 & 0.8989 & 4.1607 & 0.8678 & 7.6558 & 0.8130 & \textbf{12.7728}\rlap{*$^\dagger$} & 0.7412 & 18.2956 & 0.7084 & 20.4696 \\
\SelNet & 0.9188 & 0.9417 & 2.6518 & 0.9268 & 4.2490 & 0.8901 & 7.7238 & 0.8274 & 12.6474 & \textbf{0.7497}\rlap{*} & 17.9691 & \textbf{0.7117}\rlap{*} & 20.3631  \\
\SelNet & 0.8147 & 0.8828 & 2.4717 & 0.8734 & 4.0153 & 0.8461 & 7.5360 & 0.7965 & 12.7146 & 0.7291 & \textbf{18.3331}\rlap{*$^\dagger$} & 0.6980 & \textbf{20.5291}\rlap{*$^\dagger$}  \\
	  \bottomrule
	
	\end{tabular}

	\begin{tablenotes}
	\footnotesize
        \setlength\labelsep{0pt}
		\item The performances of the best performing model are in bold. 
		* indicates the best observed performance in a metric for a given model. 
		$\dagger$ indicates that the model performs significantly better compared to \cclerank
		according to a Wilcoxon signed rank test with Bonferroni correction at 5\% significance level.
		\par
	\end{tablenotes}

\end{threeparttable}
\vspace{-5pt}
\end{table*}

%% file: tables/hyperparameters.tex
\begin{table}[h!]
\centering
\caption{Hyper-parameters used in grid search}
\label{tbl:hyperparam}
\vspace{-10pt}
  \begin{threeparttable}
  	\begin{tabular}{
          @{\hspace{5pt}}l@{\hspace{5pt}}
          @{\hspace{5pt}}c@{\hspace{5pt}}
          @{\hspace{5pt}}p{3cm}@{\hspace{5pt}}
        }
        \toprule
         Method & Hyper-parameter & Values \\
         \midrule
         \multirow{4}{*}{\cclerank} & $l$       &  5, 10, 25, 50 \\
         					& $\alpha$ & 0, 0.05, 0.1, 0.5, 1 \\
         					& $\beta$  &  0.1, 1 \\
         					& $\gamma$ & 0, 1, 10, 100\\
       	 \hline
         \deepcdr & $l$ & 25, 50, 100 \\
         \hline
         \multirow{2}{*}{\DrugE} & hidden\_units & 128, 256 \\
         &  $M$  & 25, 50, 100\\
        \bottomrule 
       	\end{tabular}

 	\begin{tablenotes}
	\footnotesize
       	\setlength\labelsep{0pt}
		\item $l$ denotes the dimension of the latent embeddings for \cclerank and \deepcdr; 
		$\alpha$, $\beta$ and $\gamma$
		for \cclerank refer to the original notations as used in \cite{He2018}; 
		\par
	\end{tablenotes}

\end{threeparttable}

\end{table}